\documentclass[11pt]{article}

\usepackage[final]{acl}

\usepackage{times}
\usepackage{latexsym}

\usepackage[T1]{fontenc}

\usepackage[utf8]{inputenc}

\usepackage{microtype}

\usepackage{inconsolata}

\usepackage{graphicx}
\usepackage{amsmath}
\usepackage{amsthm, amssymb}
\usepackage{booktabs}
\usepackage{subcaption}
\usepackage{booktabs}

\newcommand\blfootnote[1]{%
  \begingroup
  \renewcommand\thefootnote{}\footnote{#1}%
  \addtocounter{footnote}{-1}%
  \endgroup
}

\definecolor{darkgreen}{HTML}{005e19}
\definecolor{darkblue}{HTML}{240394}
\newcommand{\code}[1]{\texttt{#1}}
\newcommand{\exampleg}[1]{\textcolor{darkgreen}{\textbf{\small{\code{#1}}}}}

\title{Some Tokens Behave like Magnets: Revealing Linguistic Organization in the Layers of Language Models}

\author{
  Andrew Liu$^*$, \quad Devan Srinivasan$^*$, \quad Gerald Penn \\
  University of Toronto \\
  Department of Computer Science \\
  \texttt{\{aliu, devan, gpenn\}@cs.toronto.edu}
}

\begin{document}
\maketitle
\blfootnote{$^*$Equal contribution.}
\begin{abstract}

We identify a special group of token vectors inside large language models (LLMs) --- which we term \emph{magnetic vectors} --- that organize the surrounding tokens by either attracting or repelling them. Particularly, tokens pointing the same way as an \emph{attracting} magnet are elongated; tokens pointing the same way as a \emph{repelling} magnet are compressed. Just as physical magnets pull or push away the iron filings around them, these vectors organize their surroundings through two opposing polarities. Moreover, we identify a statistically significant pattern in linguistic category where function words consistently act as repelling magnets in early layers, and we also find magnets consistently reorganize their polarities in unique ways deeper in the model. In a further case study we find this observation may unveil a deliberate, layer-wise organization in how LLMs process language.

This pattern is consistent across different LLM architectures, sizes, and layer configurations. It is also causally relevant. When the LLM is fine-tuned for a downstream task, the task-functional tokens emerge as magnets. E.g., in question answering, the answer-span tokens become uniquely repelling magnets in the final layer, geometrically carving the answer out of the surrounding context. Furthermore, removing early-layer repelling magnets devastates syntactic tasks (POS tagging accuracy drops from 91\% to below 10\%) while sparing semantic ones, and removing late-layer attracting magnets does the reverse. We believe this phenomenon warrants further investigation, as it opens the first probe-free path to understanding how language models geometrically organize linguistic computation across their layers.\footnote{Our code is available at \url{https://github.com/Andrwyl/SomeTokensBehaveLikeMagnets}}

\end{abstract}

\section{Introduction}

A recent thread of research on explaining black-box language models, known as mechanistic interpretability, has utilized ideas from representational geometry, where researchers analyse the geometry of vector embeddings of meaning. While fruitful, these results focus on angles and cosines, commonly ignoring lengths/norms (\citet{lee2025sharedgloballocalgeometry, elhage2022toymodelssuperposition, translatingemb2emb, ethayarajh-2019-contextual, mikolov2013linguistic} are some examples).  This seems to have happened without forethought; indeed, we surmise that most researchers in this area would readily acknowledge that both are important to a vector representations.

\begin{figure}
  \includegraphics[width=\columnwidth]{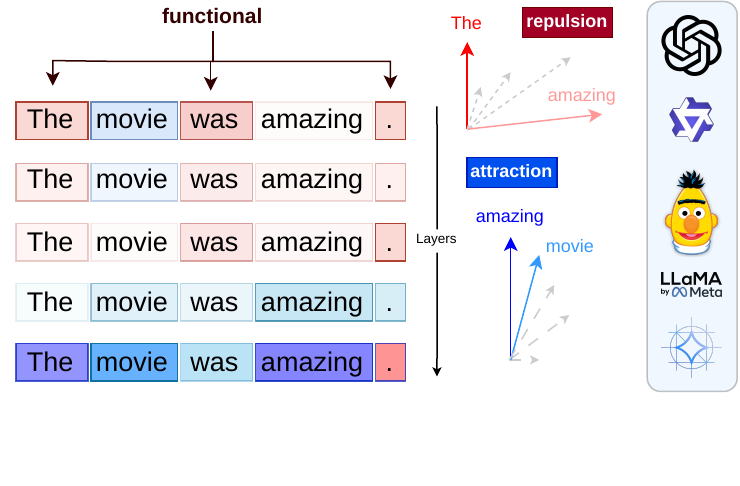}
  \vspace{-4em}
  \caption{\textbf{Magnets.} In the sentence \exampleg{The movie was amazing .} we see magnetism of different kinds as progress through the model. Early on, function tokens \exampleg{The}, \exampleg{was}, and \exampleg{.} are strong repelling magnets, shortening surrounding vectors which point in the same direction. Later on we see \exampleg{The} and \exampleg{amazing} are strong attractors, lengthening surrounding vectors that point similarly to them. The right panel shows the underlying geometry: a repelling magnet (e.g., \exampleg{The}) compresses tokens aligned with it, while an attracting magnet (e.g., \exampleg{amazing}) elongates them. This pattern is consistent across LLM families.}
  \label{fig:fig1}
\end{figure}

In this paper, we elucidate a peculiar, coarse but nevertheless linguistically interpretable geometric organization in token vector representations within stacked transformer-based language models. At some transformer layer, certain vectors serve as \textit{magnets} of a repelling or attracting polarity. Repelling magnets have the effect where vectors that point in a similar direction are shortened, and those that are more orthogonal are lengthened. Attracting magnets perform oppositely, vectors that point similarly are lengthened, while those more orthogonal are shortened. Formally, their angular parallelism to other vectors is tightly correlated with the norms of those other vectors, mediated by the depth of the representation within the transformer. We coin these curious vectors as \textit{magnets} (\autoref{fig:fig1}). Magnets exhibit clear patterns across transformer blocks and they have important consequences for downstream tasks.

Magnets naturally reveal layer-wise linguistic patterns not just in classical models like BERT, but also larger more recent models, and consistently across numerous natural language topics (healthcare, news, etc.). Magnets exhibit a progression from early to middle to late blocks of these models in a way that is consistent across different LLM families and sizes.  Even more strikingly, they discriminate between tokens based on whether the tokens are function words/affixes (determiners, conjunctions etc.) or content words/affixes (nouns, verbs etc.). But these regularities are only observable by considering both angles and magnitude.  They're found purely from the intrinsic geometry of the model, absent of any confounding auxiliary mechanisms such as probes. As such, magnetic vectors provide a new lens through which we can assess the potential functional importance of layers in neural language models.  This is, to our knowledge, the first direct, observational evidence of linguistic organization across layers that is intrinsic to the models themselves.

We additionally investigate the role of these magnets at differing layers on downstream tasks, by focusing on BERT as a case-study: if magnets support the idea of early-layer syntax and late-layer semantics, early-layer magnets should be important to syntactic tasks and later-layer magnets to semantic ones. Shockingly, we find that this is the case: early-layer magnets have outsized influence on part-of-speech tagging (a syntactic task) performance, while late-layer magnets are sharply connected with sentiment analysis and question answering (semantic tasks).

Together, our results present the first model-faithful support for layer-wise linguistic attribution, an idea long supported but never adequately substantiated, and extends it to modern large-scale language models. Magnet vectors offer a new lens and geometric perspective to understanding representations. Cumulatively, the contributions of this work are as follows:

\begin{enumerate}
    \item We identify \textit{magnetic vectors} in the representations of numerous large language models
    \item We find magnets exhibit consistent organization across layers and are sensitive to linguistic categories across models.
    \item We provide evidence that BERT's early-layer and late-layer magnets play an important role in a chosen pair of syntactic and semantic tasks, respectively.
\end{enumerate}

\section{Related Work}
\label{rw}

\paragraph{Interpretability} Our work is related to the ever-growing interest in understanding how language models work, known as \textit{interpretability}. Current popular methods in the subfield include but are not limited to attributing facets of knowledge to components of the model \cite{dai-etal-2022-knowledge, niu2024what}, attributing performance to certain salient training data points \cite{grosse2023studying, liu-penn-2025-similarity}, finding latent features that correspond to interpretable features \cite{cunningham2023sparseautoencodershighlyinterpretable, templeton2024scaling}, or leveraging topology to identify key structure in models \cite{rieck2018neural}. There has also been some success in identifying interpretable circuits and computation graphs within the model \cite{wang2023interpretability, NEURIPS2023_34e1dbe9, yu-etal-2025-sheaf}.

\paragraph{Layer-wise Linguistic Attribution} Early attempts to understand the mechanisms of BERT relied on probing, where an auxiliary model is trained to extract features from representations \cite{hewitt-manning-2019-structural}. This resulted in the proposal that the layers of BERT encode the \textit{classical NLP pipeline}, where early layers process syntax and later layers process semantics \cite{jawahar-etal-2019-bert, tenney-etal-2019-bert}. While initially embraced, this idea was heavily contested \cite{niu-etal-2022-bert} due to the use of probing: the core issue being it is unclear whether the probe itself has simply learned the property being probed for. As such, this and related work in layer-wise attribution have largely fallen to the wayside as previous methods could not disentangle their results. In addition, this line of work has rarely been expanded beyond BERT.

\paragraph{Representation Geometry}

Our finding relates to the growing body of geometric analysis within language models. This includes learned transformations that reveal striking geometric structure in days of the week \cite{engels2025not}, as well as spatiotemporal data like landmarks \cite{gurnee2024language}. These relate to the general linear representation hypothesis \cite{park2024linearrepresentationhypothesisgeometry} which posits features are stored as directions, and similar analysis reveals influential directions that represent concepts \cite{park2025categorygeometry}. \citet{lee2025sharedgloballocalgeometry} even found shared angular structure between language model spaces. These geometric approaches seldom incorporate vector length in their analysis, however. An older but highly important work is by \citet{ethayarajh-2019-contextual} which demonstrates increasing anisotropy across layers of transformer models. While their work is one of the first to show vector geometry transforming across layers, our phenomenon is unrelated to the degree of anisotropy at each layer and thus presents an entirely different geometric phenomenon that changes moving through layers.

Building upon these works, we analyse the geometry of hidden representations, but include magnitude along with angles. We find interpretable, domain-agnostic, linguistic structure across the model, crucially without auxiliary training or modules. This provides entirely model-faithful interpretability of hidden state representations.

\section{Magnetic Vectors}
\label{definition}

Magnetic vectors affect the surrounding space: Repelling vectors shorten vectors who point in a similar direction (small angle) and lengthen those far away, whereas attracting vectors do the opposite, lengthen those close in angular proximity, and shorten those far away. It's like a push and pull, repulsion and attraction, and we can mathematically formalize this phenomenon. Given input tokens $X = [x_1,x_2,x_3,\ldots,x_n]$, let $V^{l} = [v_1^{l}, v_2^{l},\ldots,v_n^{l}]$ be the vector representations for each token at layer $l$.

We can formally define vector magnetism at some layer $l$ as follows. Let $\mathrm{cos\_sim}_u:V^l\to [-1,1]$ be the cosine similarity function given as
\[
\mathrm{cos\_sim}_u(v) = \cos(\theta_{u, v})
\]

\noindent where $\theta_{u,v}$ denotes the angle between any $u, v$. Denote the norm function as $||\cdot||$. We say $u\in V^l$ is a \textit{magnet} at layer $l$ if $\mathrm{cos\_sim}_u$ is predictive of $||\cdot||$ across all tokens in the input. To determine this, we fit a linear regression between the two series: $[\mathrm{cos\_sim}_u(v):v\in V^l]$ and $[ ||v||:v\in V^l]$, and take the $R^2$ value to be the \textit{magnet strength}. The sign of the regression we define as the \textit{polarity} of the magnet. If the signed $R^2$ is negative then the magnet is repelling, and if positive it is attracting.\footnote{We use terms \emph{repelling vector}, \emph{repelling magnet}, or \emph{repelling token} interchangeably for the ease of narration.}

\begin{figure}
  \includegraphics[width=\columnwidth]{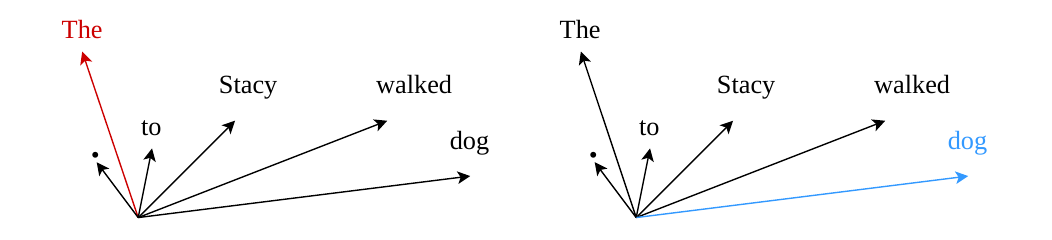}
  \caption{Toy example of \exampleg{The dog walked to Stacy} in $\mathbb{R}^2$. For repelling vectors closer angular proximity indicates a smaller norm. For attracting vectors closer angular proximity indicates a larger norm.}
  \label{fig:bert-magnet_demo}
\end{figure}

Working through the example in \autoref{fig:bert-magnet_demo}, consider the sequence of tokens \exampleg{The dog walked to Stacy}. We pass it through some model and get the outputs after layer $l$. We denote the vector representations outputted by this layer as $v_1, v_2, v_3, v_4, v_5$ where $v_5$ is the representation for \exampleg{Stacy} and $v_1$ is the representation for \exampleg{The}. Consider $v_1$ as a candidate magnet. We take the cosine similarity between $v_1$ and all other vectors, which gives us the series $A = [ \cos(\theta_{v_1, v_1}), \cos(\theta_{v_1, v_2}), \cdots, \cos(\theta_{v_1, v_5})]$. Then we compute the norms, that is $B = [ ||v_1||, ||v_2||, \cdots, ||v_5||]$. We then fit a linear regression between $A$ and $B$. Suppose the $R^2$ of the linear regression is 0.8 and the relationship is negative, we would say that at this layer, $v_1$ or \exampleg{The} is a very strong repelling magnet (strength of -0.8).

Again, geometrically the intuition is simple.\exampleg{The} is a repelling vector so other tokens with similar directions will be small in length, while \exampleg{dog} is an attracting one, so other tokens with similar directions will be long. In practice our approach can be parallelized for efficiency on very large models and samples, see \autoref{appdx:mag_defn} for details.

\subsection{Significance of Existence} \label{importance}

As it stands, it is curious to find such a phenomenon in language models, as there is no clear bias in the transformer architecture or language model training that would encourage such organization. We find every model frequently exhibits magnets as strong as $\pm0.7$ to $\pm0.8$ regardless of context which is a very significant relationship, indicating this organization is not noise or some anomaly. Further investigation, presented in detail later, reveals these magnets follow layer-wise patterns aligned across various model architectures, and exhibit clear linguistic separation and impact on downstream tasks which sediments the impact of this discovery.

\section{Experiments}
Our main line of experimentation investigates the patterns that magnets follow across different inputs and models. The core questions we seek to answer are whether magnets emerge in certain ways at different points in the model (early, middle, late) and whether certain types of tokens tend to be attracting or repelling vectors. We find that, extraordinarily, magnets have linguistically-guided layer-wise and token-wise patterns that are stable and consistent across many contexts and models.

\begin{figure*}
\centering
  \includegraphics[width=\textwidth]{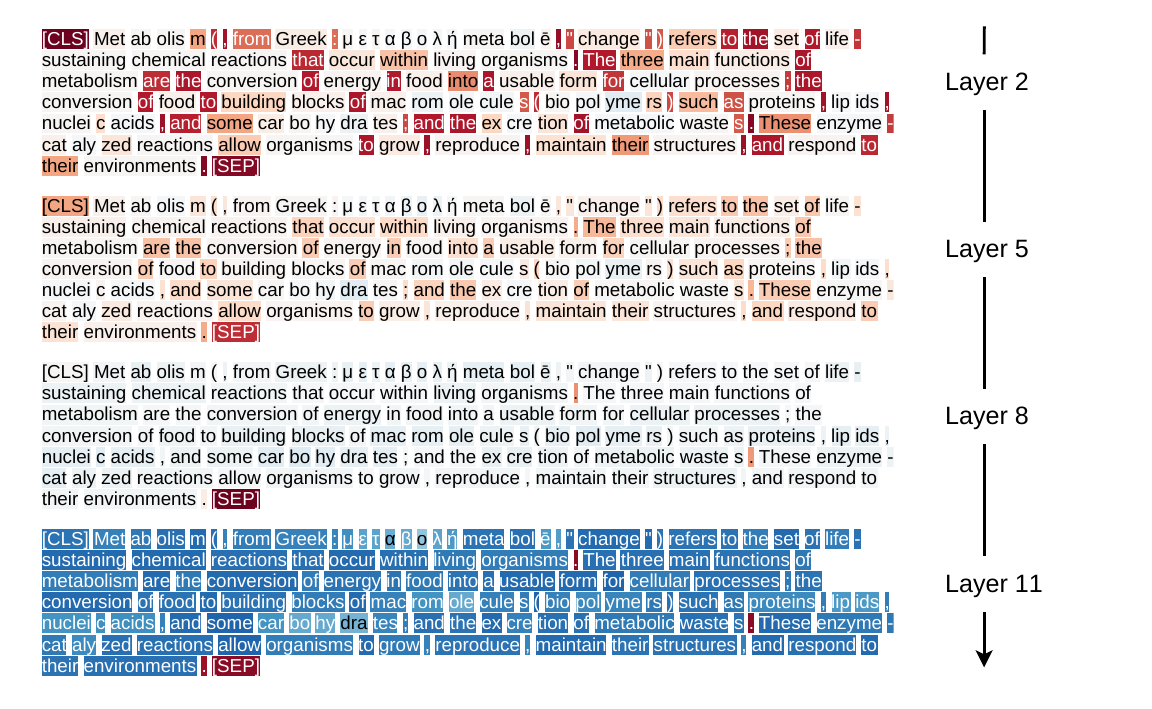}
  \caption{The evolution of magnet tokens moving through BERT. Early layers exhibit only repelling magnetic vectors (\textbf{\color{red} red}) and they tend to be function tokens, moving through the layers they change gradually in intensity and polarity until strongly attracting magnets (\textbf{\color{blue} blue}) become the majority.}
  \label{fig:bert-output}
\end{figure*}

\subsection{Setup}
\label{main:setup}

We manually curate a dataset of 100 paragraphs, each around 200 words, spanning different contexts like news articles, healthcare sites etc (details in \autoref{appdx:dataset}). We investigate five models, BERT \cite{devlin-etal-2019-bert}, GPT2 \cite{radford2019language}, Qwen2.5-3B \cite{Yang2024Qwen25TR}, Gemma3-1B \cite{gemmateam2025gemma3technicalreport}, and Llama3.2-3B \cite{grattafiori2024llama3herdmodels}. These models are chosen for specific reasons: BERT due to the large amounts of previous work attempting to extract linguistic structure from its representations at a layer-level, GPT2 as a foundational, well-studied decoder model in interpretability research, and Qwen/Gemma/Llama to validate that our structure exists on larger-scale modern models as well. For each paragraph, we observe the magnet properties of each token at each layer for each model.

These experiments are motivated by a surprising observation: language models tend to delegate function tokens as repelling magnets early in the model, one of the first signs of linguistic separation being encoded natively in the representations. In addition, early layers contain magnets that are primarily one polarity, middle layers exhibit a transition period where the polarities of magnets change and possibly change back (a \textit{bowtie} effect), before establishing themselves in the later layers. For example in \autoref{fig:bert-output}: in early layers the function tokens are consistently identified as repelling (red) before a phase transition in the middle and the attracting magnets are established in the late layers. \autoref{appdx:linguistic_striation} presents visual examples for the other models, and \autoref{appdx:polarity_transition} shows this \textit{bowtie} transition phenomenon in the middle layers holds for all models across our dataset. To comprehensively test that early-layer repelling magnets are consistently function tokens, we use a part-of-speech tagger to tag a word with its part-of-speech type. If the word is labeled with a content tag (nouns, verbs etc.) we give it a label of 1, otherwise it is labeled 0. For each word in the dataset and for each layer, we now have a label for whether it is function or content as well as a magnet strength. We compute a Point-Biserial Correlation (PBC) between these two variables at each layer, allowing us to test whether there is indeed a correlation between magnet polarity and syntactic category at each layer.

We must include special consideration for our decoder models. Our magnet property is a property of a token's representation relative to the other representations in the input. Thus, there is an issue with directly applying the calculation to decoder representations. Since tokens do not see forward in the input, it is meaningless to calculate a regression score between some token property with any tokens after it, as they are independent by design. We leverage the main idea in \citet{leviathan2025prompt} to work around this, repeating the input, then only computing magnets on the repeated portion which can attend to itself via the earlier instance. Details of this implementation and other modifications we had tried are explained in \autoref{appdx:decoder}.

\begin{figure}[t!]
  \includegraphics[width=\columnwidth]{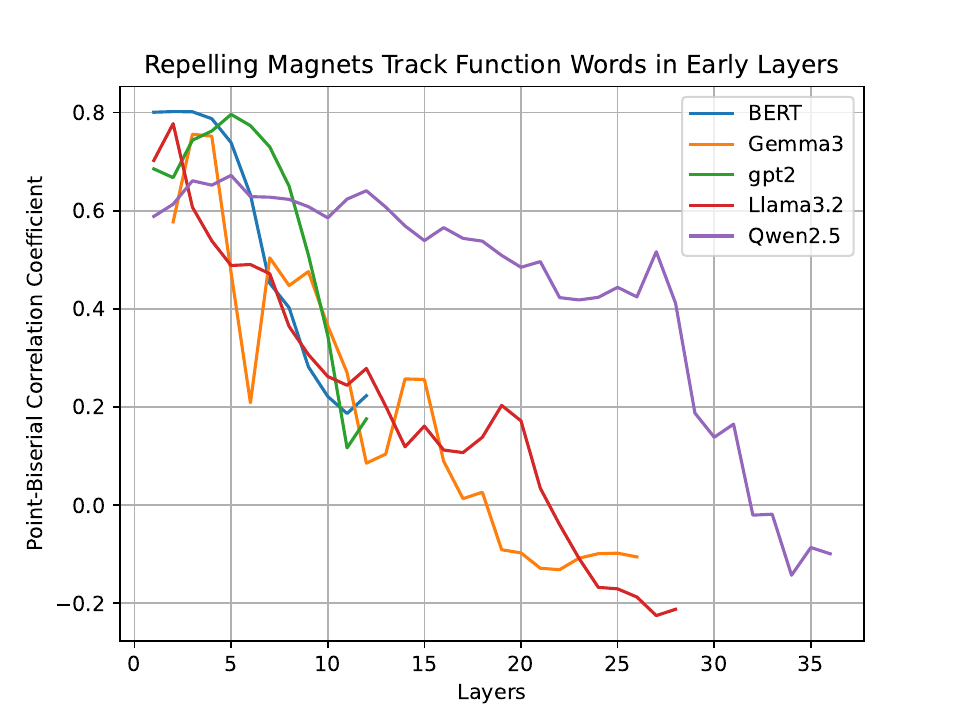}
  \caption{Correlation between magnet strengths and linguistic token type (function or content) at each layer. The early-layer \textbf{\color{red} repelling vectors} are very often function tokens and this tapers off moving deeper in the models.}
  \label{fig:all-striations}
\end{figure}

\subsection{Results}
\label{main:results}

\autoref{fig:all-striations} presents the point-biserial correlation coefficient for each model across their layers. We see that our initial observations are corroborated. At the early layers for the models, the correlation coefficient is as high as 0.8 indicating a very strong connection between the magnet properties and the syntactic type of a token. Function tokens \textit{overwhelmingly} tend to be repelling magnets at the early layers. Moving deeper in the model, the coefficient begins to decline in-line with that \textit{bowtie} transition phase we previously observed before late-layer magnets are formed.

The establishment of this layer-wise structure emerging across models hearkens back to previous work connecting BERT to the classical NLP pipeline, which roughly proposed that early layers processed syntax while later layers processed semantics. We can see that our magnet structure has striking similarities to the proposed pipeline due to its layer-wise nature. The pipeline proposes a syntax regime in the early layers, and it is in those early layers we observe repelling magnet structure separating syntactic and semantic tokens. The pipeline proposes a semantics regime in the late layers, and we find those late layers are the product of the \textit{bowtie} transition of magnet polarities. The connection is clear: our magnet structure indicates syntax early, a transition in the middle, and hints at semantics emerging late. We explore this deeply in subsequent experiments.

We have obtained a natural way of proposing a projection of layers of language models onto the pipeline that sidesteps criticisms leveled at previous attempts to do so: we leverage only the intrinsic geometry of the representations. In addition, we have shown these linguistically-guided layer properties exist not just for BERT, but in decoder models like GPT2 as well as modern, large-scale models like Qwen, Gemma, and Llama.

\section{BERT and Downstream Tasks}
\label{downstream}
We wish to obtain a stronger understanding of the role these identified magnets play in the processing steps of language models. If early-layer magnets are indeed tied to syntax, they should be important for syntactic tasks. If late-layer magnets are tied to semantics, they should be important for semantic tasks. For these experiments we focus on BERT for two reasons: first, it is the model primarily studied in the context of layer-wise attribution, making it a natural point of comparison, and second, as an encoder model its tasks are more closed-ended and controlled. Tasks like part-of-speech tagging and sentiment classification have clearly defined syntactic and semantic interpretations allowing us to cleanly test the layer-wise linguistic categories we identify. In contrast, decoder models are more naturally suited for tasks such as open-ended reasoning which is harder to decompose to the level of linguistic granularity that we would like. Our focus on BERT allows us to draw the most direct conclusions about our hypothesis, and is most grounded in previous work in the field. We select the following tasks for our study:

\paragraph{(Syntax) Part-of-Speech Tagging (POS)} This tasks a model with classifying the grammatical category of each token in the input sequence. Ex: \exampleg{The} $\to$ determiner (DET), \exampleg{Sword} $\to$ verb (VB). This is a highly syntactic task as it requires sensitivity to the grammatical role each token plays in the sentence structure rather than its meaning.

\paragraph{(Semantics) Sentiment Analysis} This tasks a model with classifying the sentiment of a sample of text as positive or negative. Ex: \exampleg{The play was terrific} $\to$ Good. This is a highly semantic task as it requires sensitivity to the meaning of the words rather than just grammatical structure.

\paragraph{(Semantics) Question Answering (QA)} This tasks a model with identifying the answer to a question in some text. Given the question and some context, it must identify the answer. Ex: \exampleg{What colour is the sky? The sky is blue and clouds are white} $\to$ \exampleg{blue}. This is another highly semantic task, as meaning sensitivity is of key importance.

\begin{figure*}[th!]
\centering
  \includegraphics[width=\textwidth]{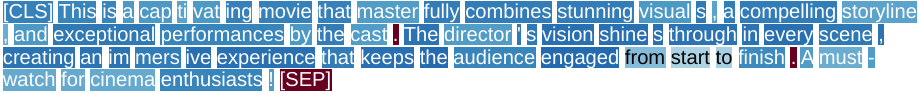}
  \caption{Sentiment Analysis at Layer 10: \textbf{\color{blue} Attracting} tokens are substrings of high sentiment, and in this figure the darkness of the colors are key, not just polarity. Adjectives like \exampleg{stunning}, \exampleg{compelling}, and \exampleg{exceptional} as well as \exampleg{creating an immersive experience} are noticeably darker, indicating they are also more attracting.}
  \label{fig:bert-sst-output}
\end{figure*}

\subsection{Setup}
We fine-tune BERT on three standard datasets: CoNLL-2003 \cite{tjong-kim-sang-de-meulder-2003-introduction} for POS tagging, SST2 \cite{socher-etal-2013-recursive} for sentiment analysis, and SQuAD v1 \cite{rajpurkar-etal-2016-squad} for QA, yielding $3$ models. In our main experiments we found magnets identified function tokens in the early layers, which we interpreted as a correspondence with the proposed syntactic processing regime of previous work. Additionally, we identified that when moving deeper into the model this linguistic separation fades away, and the model exhibits a transition period re-orienting the polarity of the majority magnets.

We hypothesized that this transition aligned with the proposed transition from syntactic to semantic processing. In fact our preliminary investigation with BERT revealed extraordinary connections between late-layer magnetic patterns and the sentiment classification and QA tasks (our semantic tasks): In sentiment analysis, late-layer, strong, attracting magnets were tokens most ostensibly related to the sample's sentiment (see \autoref{fig:bert-sst-output}). When fine-tuned for QA, at these same late layers we observed that the answer spans are among the only strong repelling magnets whereas every other token is attracting, so magnets effectively carve the answer out of the paragraph (see \autoref{fig:bert-qa-output}). \autoref{appdx:downstream} contains additional examples of the sentiment and QA task inputs.

Motivated by these observations we wish to exhaustively test the impact of the early and late-layer magnets on syntactic and semantic tasks, through two inquisitions. First we causally ablate magnets at the early and late layers by zeroing out (replacing with the zero vector) strong repelling magnets early on, and strong attracting magnets later on (strong is $R^2 > 0.5$), and report the effects on syntax (through POS tagging) and semantics (through sentiment analysis). We also include the baseline of separately ablating an equal number of random tokens at each intervention. Note we do not ablate in middle layers as there are so few strong magnets to begin with.

The second inquisition is QA, where in particular we know the answer tokens, so we can directly test the alignment between answer spans and magnetic polarity. We do so with the Wilcoxon Signed Rank test, and report the Rank Biserial Correlation to quantify the effect size. This effectively tests if answer tokens are more repelling than non-answer tokens and by how much, and serves to further explore the role of late-layer semantics in the model.

While we experiment with fine-tuned (not base) models, we can understand how magnet geometry influences specific linguistic tasks, and speculate at their function, which is undoubtedly useful. For more details on this setup see \autoref{appdx:downstream}.

\subsection{Findings}
\label{downstream:results}

\autoref{tab:ablating_early} and \autoref{tab:ablating_late} show the results of our first line of experiments: causal ablations. We report the classification accuracy on the CoNLL test set for POS tagging, and SST2 test set for sentiment analysis. In \autoref{tab:ablating_early} we see that at these early layers (precisely where the PBC was strongest), ablating the repelling tokens has catastrophic effects on syntax performance (POS), further solidified when compared to the random baseline which yields a comparatively mild impact, despite zeroing-out an equal number of tokens. For reference, the unmodified accuracy is 91.5\%. In comparison, when ablating these early magnets for sentiment classification, a semantic task, the performance impact is much weaker. In this task the unmodified accuracy is 94.9\%. Thus, these early magnets appear to be specially important for syntactic tasks. In \autoref{tab:ablating_late} we see the other half, ablating late-layer attracting vectors produces severe drops in semantic performance while the baseline is again much more mild. As expected, ablating these late magnets for POS tagging, a syntactic task, produces negligible performance degradation. This affirms what we hypothesized, that late magnets are especially important for semantics.

\begin{figure*}[th!]
\centering
  \includegraphics[width=\textwidth]{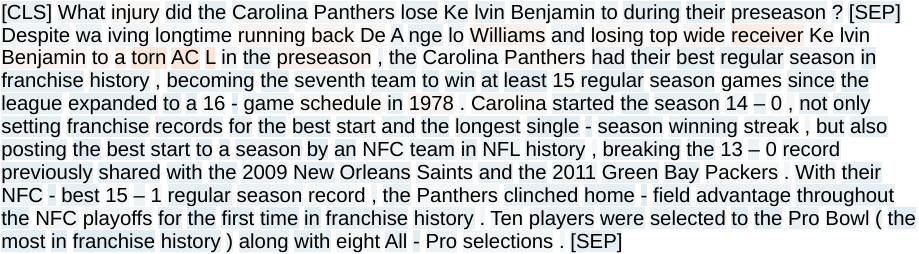}
  \caption{Question Answering at Layer 12: We see the answer tokens (\exampleg{torn ACL}) are the strongest \textbf{\color{red} repelling} tokens amongst mainly \textbf{\color{blue} attracting} magnets}
  \label{fig:bert-qa-output}
\end{figure*}

In \autoref{tbl:bertQA} we report our findings for our second experiment line on the SQuAD validation set. It corroborates what we had initially found (\autoref{fig:bert-qa-output}): late-layer answer tokens are uniquely delegated as more repelling. This indicates that magnetism clearly separates the answers from the paragraph for question answering, impressive given that answer tokens represent less than $3\%$ of all tokens. We also see the pattern peaks at the final layer where we saw that answer tokens were entirely repelling magnets. Accumulating all our results, these findings show that in BERT early-layer magnets are crucial for syntactic tasks while late-layer magnets are important for semantic ones, in-line with prior proposed stages of linguistic processing.

\begin{table}[h]
\centering
\resizebox{\columnwidth}{!}{
\begin{tabular}{c|cc|cc}
\toprule
\textbf{Layer} &
\multicolumn{2}{c|}{\textbf{Sentiment Classification}} &
\multicolumn{2}{c}{\textbf{POS Tagging}} \\

\cmidrule(lr){2-3}
\cmidrule(lr){4-5}

& Random & Magnet & Random & Magnet \\
\midrule
1              & 0.7792                     & 0.6111                    & 0.7398              & 0.0679             \\
2             & 0.8375                     & 0.8083                    & 0.7852              & 0.1132             \\
3             & 0.8958                     & 0.8181                    & 0.8577              & 0.4274             \\
\bottomrule
\end{tabular}}
\caption{\textbf{Early-Layer Repelling Magnets Matter for Syntax} Effect of zero-ablating an equal number of random vectors and early-layer repelling vectors on classification accuracy of a syntactic (POS) and semantic (sentiment) task. Early-layer magnets are far more important to the syntactic task.}
\label{tab:ablating_early}
\end{table}

\begin{table}[h]
\centering
\resizebox{\columnwidth}{!}{
\begin{tabular}{c|cc|cc}
\toprule
\textbf{Layer} &
\multicolumn{2}{c|}{\textbf{Sentiment Classification}} &
\multicolumn{2}{c}{\textbf{POS Tagging}} \\

\cmidrule(lr){2-3}
\cmidrule(lr){4-5}

& Random & Magnet & Random & Magnet \\
\midrule
10              & 0.9167                     & 0.8514                    & 0.9490              & 0.9490             \\
11             & 0.9097                     & 0.7625                    & 0.9490              & 0.9490             \\
12             & 0.9042                     & 0.6958                    & 0.9291              & 0.9015             \\
\bottomrule

\end{tabular}}
\caption{\textbf{Late-Layer Attracting Magnets Matter for Semantics.} The effects of zero-ablating an equal number of random vectors and late-layer attracting vectors on classification accuracy of a syntactic (POS) and semantic (sentiment) task. These late-layer magnets are far more important to the semantic task.}
\label{tab:ablating_late}
\end{table}

\begin{table}[h]
\centering
\resizebox{0.8\columnwidth}{!}{
\begin{tabular}{c|c|c|c}
\toprule
\textbf{Layer} & \textbf{Effect Size} & \textbf{RBC} & \textbf{$p-$value} \\
\midrule
$1-8$ & $-$ & $-$ & $>0.05$ \\
\midrule

$9$ & $1.1197\times10^7$ & $-0.5987$ & $<10^{-53} $\\
$10$ & $0.1717\times10^7$ & $-0.9385$ & $<10^{-53} $\\
$11$ & $0.1806\times10^7$ & $-0.9353$ & $<10^{-53} $\\
$12$ & $0.0028\times10^7$ & $-0.9990$ & $<10^{-53}$ \\
\bottomrule
\end{tabular}}
\caption{\textbf{QA Magnets Carve Answer Spans.} We present the Wilcoxon Signed Rank Test results (lower effect is stronger). In late layers we see extremely significant correlation, indicating answer tokens are much more repelling than non-answer tokens.}
\label{tbl:bertQA}
\end{table}

\section{Discussion}

\paragraph{Core Findings} Our main findings in \autoref{main:results} present some critical takeaways. The first is that magnetic vectors reflect a geometric organization of the representations within language models, one that evolves and transforms through the model (\autoref{appdx:polarity_transition}) and notably one that is deliberate (see \autoref{importance}). The existence of strong magnetism and their consistent patterns across models directly indicates that this geometric structure is likely inherent to transformer language models themselves, and is one of few methods to analyze through vector magnitude alongside angular geometry. Secondly, the fact that magnet strengths and linguistic type exhibit very high PBC scores early on (see \autoref{fig:all-striations}) indicates that the structure is not only deliberate, but reasonably governed by linguistic principles. Finding interpretable patterns inside language models is itself difficult especially purely based on hidden space geometry, so this is quite striking. Then there is the connection to prior work, as magnetism presents us with a new way of revisiting the classical NLP pipeline projected onto BERT, where the model abstractly begins with syntax early on then transitions to semantics later. As mentioned we see clear ties to syntactic processing at the early layers of a model (\autoref{tab:ablating_early}), followed by a \textit{bowtie} transition where the model reorganizes vectors (\autoref{appdx:polarity_transition}), then the end where new magnets are established (\autoref{fig:bert-output}) with less syntactic separation. Although the late-layer magnets are homogeneous and don't exhibit the function/content striation, this does not disparage the potential of using magnetism to interpret the model at this stage as they are highly related with specialized downstream tasks. We find these late-layer magnets correspond strongly to semantic tasks in one model (\autoref{tab:ablating_late}, \autoref{tbl:bertQA}). These results let us reconcile what previous probing results had found, while avoiding the criticisms they faced. Although our other layer-wise patterns extend to all model types, we cannot yet claim other models also exhibit the same strong causal links to downstream performance until they are studied in future work. But in the interim, these results suggest that through magnetism we can begin to piece together the general form of the classical NLP pipeline: layers encode linguistic properties across all models, and within BERT early magnetism is important to syntax while late magnetism is crucial to semantics.\\

\paragraph{What Causes Magnetism?} It remains to be discussed why this phenomenon arises, especially positioned relative to results in circuit discovery \cite{yu-etal-2025-sheaf}, computation graphs \cite{ameisen2025circuit} and SAE feature discovery \cite{cunningham2023sparseautoencodershighlyinterpretable}. Research \cite{elhage2021mathematical, elhage2022toymodelssuperposition} has proposed transformers may read and write to subspaces within the representation space at various times, plausibly leveraging geometric properties \cite{sun2026valencearousalsubspacellmscircular, tiblias2026hypothesisdrivenfeaturemanifoldanalysis} to do so. Contrary to the aforementioned pipeline, in their work layer-wise computation does not necessitate some iterative, "bottom-up" approach, but can be composed of disjoint, unordered points of computation. Yet based on our results we posit that general inherent qualities to language processing, like handling syntax before semantics, remain advantageous for a transformer to learn iteratively, whilst simultaneously models indeed may still compute via subspace manipulations. We believe magnetism likely occurs because of how features are delegated to token representations and moved around in these manipulations subject to some overarching processes (syntactic or semantic). Given that mechanistically transformers read and write with non-linear projections and residual updates, it's plausible that repelling magnets let the model reserve important directions (perhaps features) from others at certain steps to mitigate noise in these read/write operations (making similar directions small or pushing them away). Similarly, attracting magnets may be used for highlighting important directions which need to dominate, making similar directions large. It is still difficult to draw conclusions, but akin to the intuitive representational geometry in prior work \cite{tiblias2026hypothesisdrivenfeaturemanifoldanalysis, sun2026valencearousalsubspacellmscircular}, we believe some learned geometric organization is leveraged based on unknown features for the model's computation, and the magnetism phenomena is a remarkable consequence of this.

\section{Conclusion}

We have identified new structure in the vector representations of large language models. Experimentally, these vectors reveal layer-wise linguistic patterns across model sizes and architectures. We show causal evidence in BERT that early-layer magnets influence a chosen syntactic task more than a chosen semantic/pragmatic task, and \textit{vice versa} for late-layer magnets. Crucially, this discovery was the first to date that was made intrinsically within the model, without probing, letting us reaffirm the contested results in past work. Beyond this, magnetic vectors offer a new geometric framework for future work in understanding language models.

\section*{Limitations}

We must first highlight that our evidence of magnetism exhibiting linguistic separation applies only to English models on English tasks. Extending this to other languages is important future work to test whether this finding holds universally. Though our linguistic patterns can be found in not just BERT, but modern model families like Gemma, Qwen, and Llama, our downstream case-studies are restricted only to BERT. As mentioned in the paper, this is because BERT is so extensively cited in the layer-wise approaches that we directly align our findings with. In addition, BERT's naturally suited tasks offer us more linguistic control compared to the tasks suited for decoder models, allowing us to more accurately test our hypotheses. We chose these three tasks with BERT because they are well studied and very linguistically aligned (either predominantly syntactic or semantic). It would be perhaps more comprehensive to investigate more nuanced tasks, which we leave as an intended avenue of future work. Lastly, due to the auto-regressive nature of these large models, we needed to leverage a trick to uncover this underlying structure that as of right now causes difficulty in directly adapting our magnet ablations downstream.

However, we assert that the linguistically-guided structure indeed exists in these larger models as shown in our main line of experimentation. We view that our rigorous experiments with BERT are a necessary foundational step, well-situated in related interpretability literature, to validate this new-found phenomenon downstream. After laying the groundwork, we leave leveraging this structure on decoder models and conducting similar investigations into their downstream effects to future work.

Finally, while we connect our finding to other works in interpretability in a speculative way, we do not have a concrete explanation for why this phenomenon occurs. What we have done is comprehensively establish that this structure has undeniable ties to linguistic processing across a wide range of models. We leave an investigation of what causes these vectors to appear to future work.

\bibliography{custom}

\appendix

\section{Magnetic Vector Implementation Details}
\label{appdx:mag_defn}
While on paper identifying magnetic vectors requires performing linear regression for each token in an input sequence, this can be done in parallel as the problem reduces to a single least squares matrix equation, allowing for efficient runtime even on very large models. This implementation, and all experiments are available in our repository.

Furthermore magnetism is assessed across \textit{tokens} not words, which includes special tokens. This makes it easily adaptable to any model and tokenization architecture. 

\section{Dataset Curation}
\label{appdx:dataset}

\paragraph{Collection} We hand craft a dataset of $100$ paragraphs by sampling documents online from the following general domains: Sports, Politics, History, Medicine, News, Technology, and Science. We searched each topic online looking for well formatted paragraphs representing complete ideas. All data was collected by our research team, with no external help or annotators, and was collected from freely available online sources. 

\paragraph{Purpose} The intention was to have no bias in how this data was collected so as to honestly survey natural language across a wide variety of differing domains that compose the English language online, as this is what modern models are trained on. Recall from our experimental setup (\autoref{main:setup}) that we tag each word with it's part-of-speech category. However, magnets operate on tokens not words, so this meant that for some tokenizers, we had to manually align each token with the word, so we can infer the relevant token-level label from the base word's label. Recall the label is constructed from the part-of-speech category: 0 for function and 1 for content. This alignment was laborious, which is why our dataset is the size it is. However, given it's size and sparsity across all natural language and starkly different domains, the fact that we observe a strong unanimous pattern across model families, and model sizes, solidifies our results as an inherent property to transformer-based language modelling.

\section{Magnet Polarity Transition Across Layers} \label{appdx:polarity_transition}

\begin{figure*}[t]
    \centering

    \includegraphics[width=0.32\textwidth]{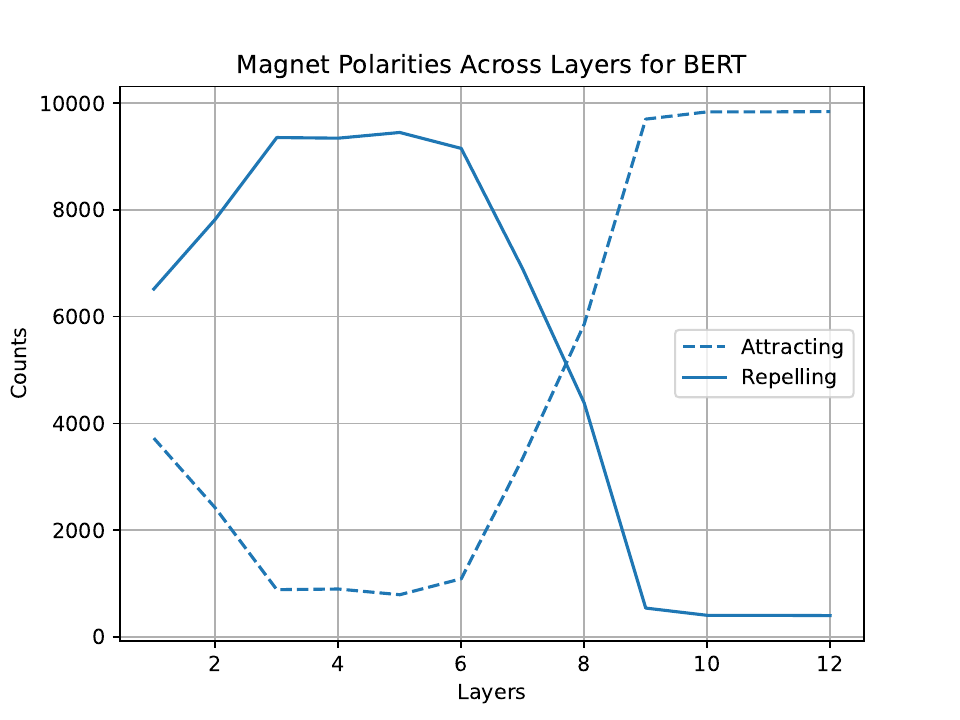}
    \hfill
    \includegraphics[width=0.32\textwidth]{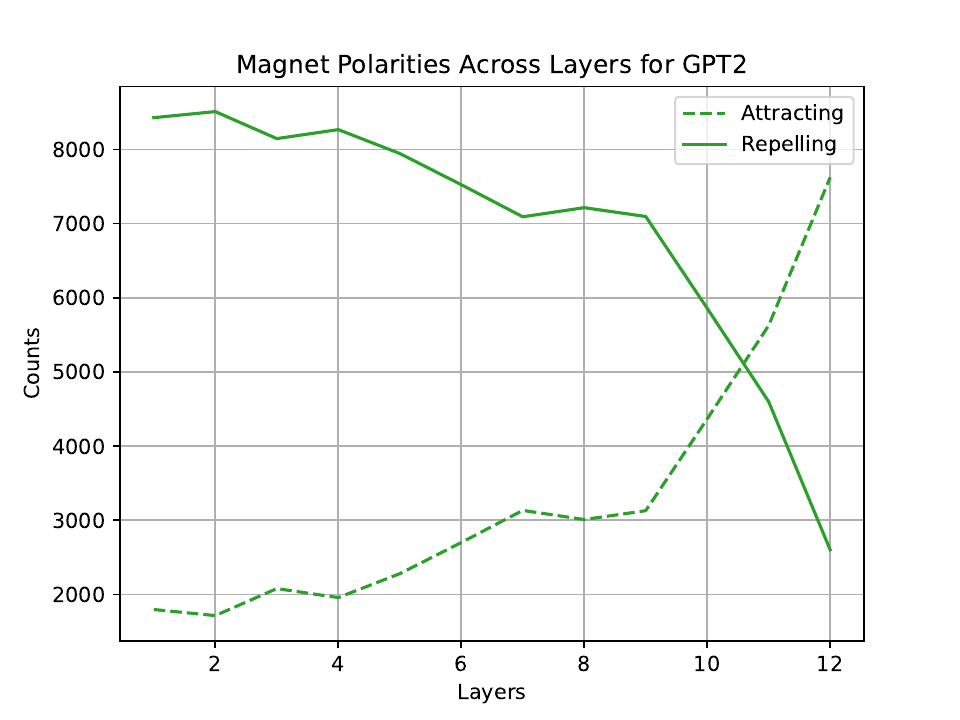}
    \hfill
    \includegraphics[width=0.32\textwidth]{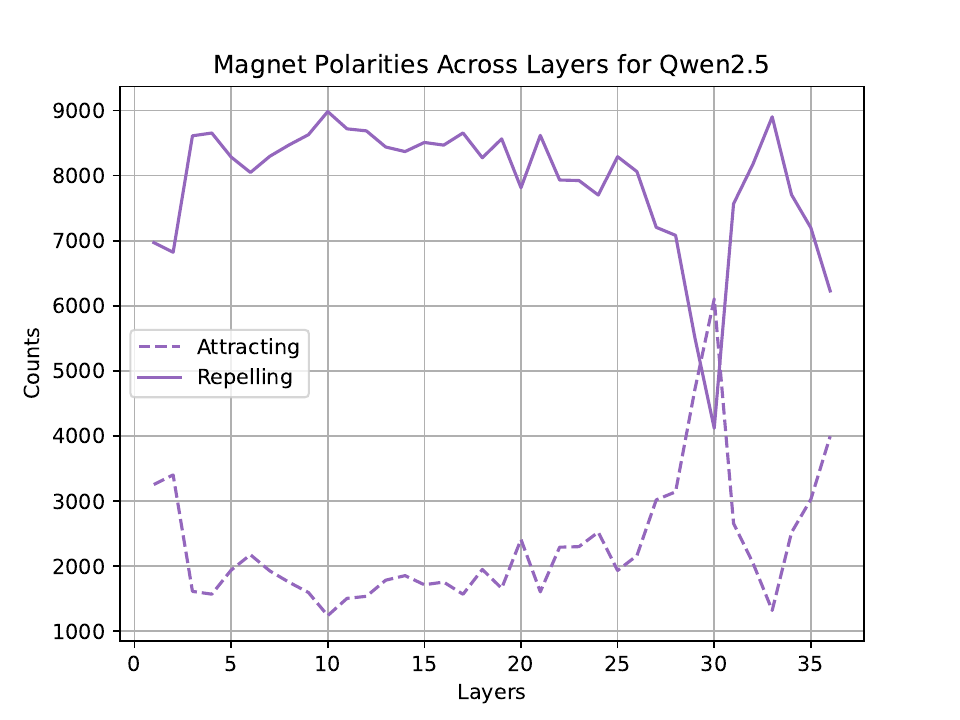}

    \vspace{0.5em}

    \makebox[\textwidth][c]{
        \includegraphics[width=0.32\textwidth]{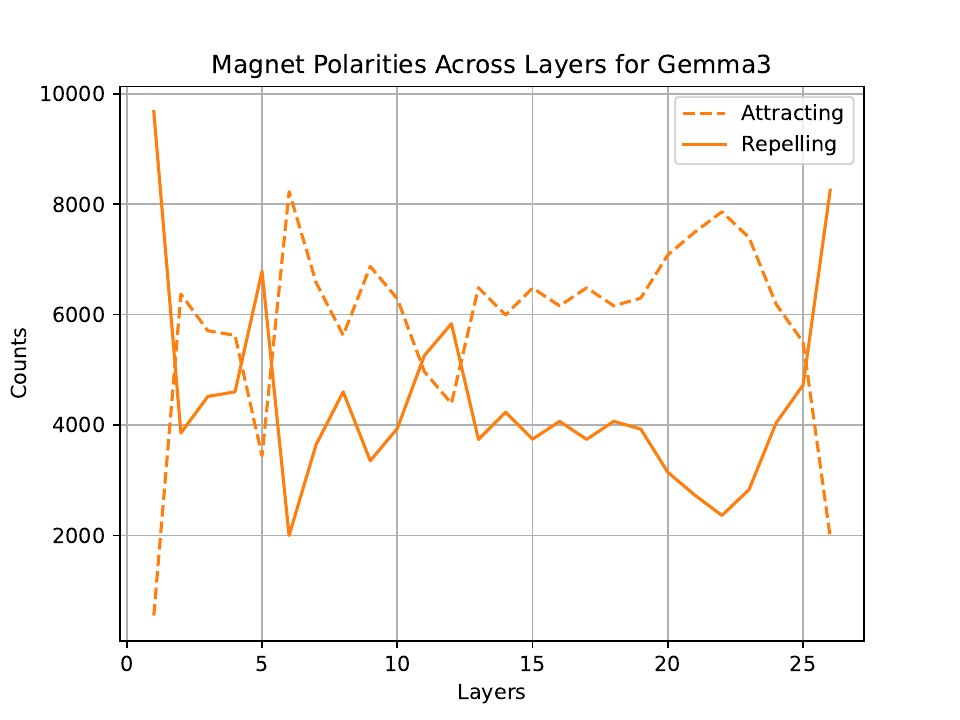}
        \hspace{0.03\textwidth}
        \includegraphics[width=0.32\textwidth]{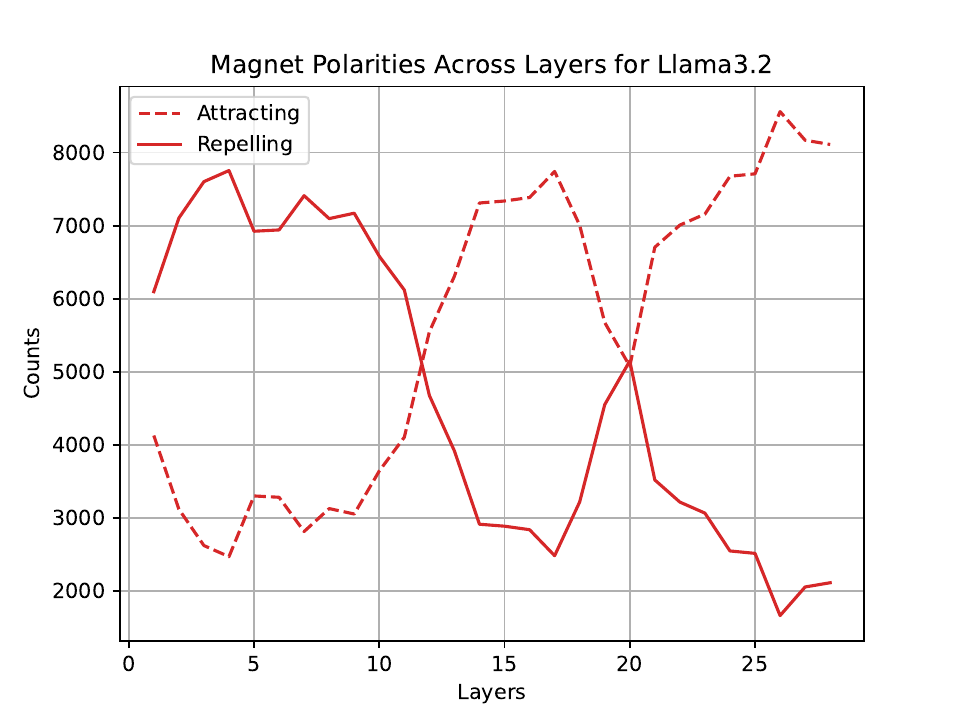}
    }

    \caption{
    Transition in magnet polarity moving across layers. The solid line represents the number of repelling vectors in the entire dataset, the dashed one represents attracting. We observe a bowtie effect phenomenon indicating a transitionary period for all models. Note this counts the number of negative and positive magnets, but does not include their strengths.
    }
    \label{fig:layer_transitions}
\end{figure*}

We present the change in the polarity counts of magnets across our dataset as we move deeper through the model. We see in \autoref{fig:layer_transitions} that for all our models, there is a clear general pattern present.

\paragraph{Early Layers} In the early layer we see a domination of repelling vectors, with some models like GPT2 and Qwen2.5 maintaining this one-sided behaviour for most of the layers. Our results in \autoref{main:results} demonstrated that at the early layers (not all layers where repelling vectors are present), the polarity of these magnets identify function and content tokens, indicating a syntactic striation. Note that the counts in these figures do not incorporate magnet strength, so indeed many of the repelling and attracting vectors may be weak.

\paragraph{Transition} In the middle of these layers we see consistent reorganization. Especially in the larger models (Qwen, Llama, and Gemma) we see transitions between the dominating polarity, and in some cases a reversion happens multiple times as with Gemma. We refer to these transitions as a \textit{bowtie} effect (based on the graph shape). However the characteristic of each \textit{bowtie} is unique to the model, and given our downstream results with BERT (\autoref{downstream:results}), this has an interesting implication. As we saw with BERT, early-layer repelling magnets, and late-layer magnets (of both polarities) had importance on specific linguistic tasks. In the case of these transitions, we see they are often gradual (not usually sporadic between adjacent layers), which suggests to us that this reorganization is deliberate, and optimistically, perhaps a reflection of some specific linguistic task at that period in the model. It is also possible that these nuanced transitions are a reflection of intricate differences in model architectures. This is all of course speculation, based on the results we observe so far. 

As mentioned in our experiments (\autoref{downstream}) we seldom see any strong magnets in the middle layers of BERT, thus we cannot use similar causal ablations to understand that transition. Furthermore, adapting our ablations to decoder models with tasks of similar linguistic granularity stand as an intended avenue for future work. 

\paragraph{Establishment} At the final layers, with the exclusion of Qwen, we see establishment of one magnet polarity. We believe this may be due to the language modelling heads (projections, classifiers, etc.) atop the transformer layers which warrants some kind of hidden space geometric stabilization before prediction. Of course this is a hypothesis, and further work is needed to better understand the behaviour.

\section{Special Consideration for Decoder Models} \label{appdx:decoder}

We present solutions we had tried to mitigate the auto-regressive problem in order to allow our magnet calculations to be applied to decoder models. One possible tweak would be to calculate the regression score backwards, that is between angles and norms for every token in the sequence prior to the candidate magnet token. This is effectively applying the causal mask on our magnet calculation. However, this suffers from a new issue, since earlier tokens have far fewer other tokens that need to be considered, which skews the regression scores in a way that is not intuitively reconcilable.

To remedy this, we use an idea from \citet{leviathan2025prompt} involving giving an input twice to a model. As such, the token representations of the second instance are imbued with information about the other tokens in the input. Specifically, for each paragraph we append it to itself before giving it to GPT2, Qwen2.5, Gemma3, and Llama3.2. We then exclusively look at the representations of the second half of the input and run our magnet calculations the normal way, no causal masking needed, ignoring the original instance in the sample. This allows us to keep balanced counts for each token while also ensuring each token has meaningfully seen the other tokens in the sequence.

\section{BERT Downstream Tasks}
\label{appdx:downstream}
\begin{figure*}[h!]
\centering
  \includegraphics[width=\textwidth]{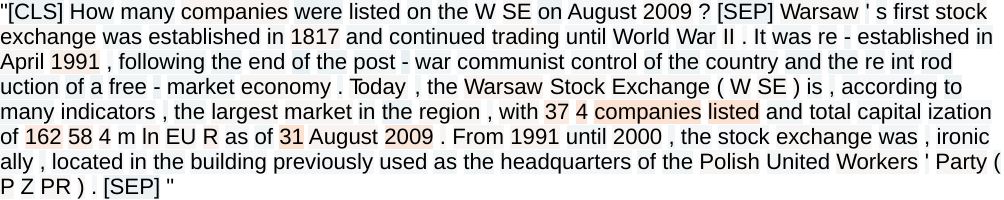}
  \caption{A layer 12 example of BERT for question answering, where the answer is $374$. Here we see clear answer carving in \exampleg{374 companies listed}, where other numbers are also considered as repelling magnets.}
  \label{fig:qa_2}
\end{figure*}

\begin{figure*}[h!]
\centering
  \includegraphics[width=\textwidth]{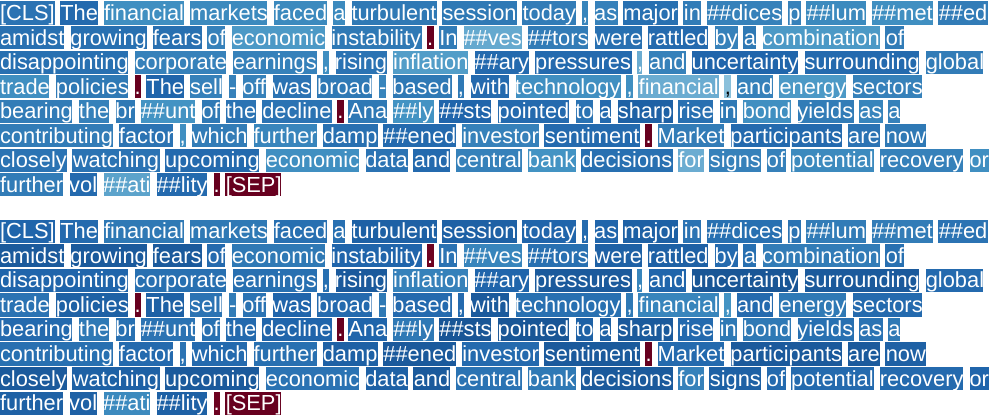}
  \caption{Layers 10 and 11 in BERT for sentiment classification, we see the white font text (corresponding to the strongest magnets) are often sentimental adjectives like \exampleg{turbulent}, \exampleg{major}, \exampleg{instability}, \exampleg{disappointing}, \exampleg{sharp rise}. Here the sentiment is negative, correctly classified by the model.}
  \label{fig:sa_2}
\end{figure*}

See \autoref{fig:qa_2} for another question answering example, and \autoref{fig:sa_2} for another sentiment analysis example. Below we will overview some technical details regarding Question Answering and the significance testing. For all tasks we train according to prior research to reach competitive accuracies. We train on a local cluster of Nvidia 24GB L4 GPUs using $2$ to $4$ depending on the task.

\paragraph{Details} Question Answering is a unique task. The model uses a classification head on top of BERT that outputs two logits, $S, E$, over the input sequence. These logits correspond to each token being the start and end of the answer span respectively. It is our job then to decode the answer from these two logit series, where in practice we use the native implementation given by HuggingFace.

\paragraph{Ablations vs. Statistical Testing} In the case of question answering, since we know the answer tokens, computing a correlation is better suited than ablations. In tasks like POS or sentiment analysis, it's unclear what tokens contribute to the final classification and how they interact with each other. For example, consider the sentiment of the sample \exampleg{The play is not good}. Does \exampleg{good} only matter? How does \exampleg{not} affect the other tokens? Or in POS, how does a tagger determine the tags for \exampleg{watch} and \exampleg{play} in \exampleg{We watch the play}, as they can be verbs or nouns. In this foggy attribution situation, causal ablations let us confidently conclude that magnetism is relevant. However with QA, we observed that magnetism carves out the answers, so if we zero-ablate those tokens, was it the magnetism that mattered or trivially the tokens themselves, because obviously the model cannot answer the question without the answer context. These results would be inconclusive. But, the fact that it's carved out is what's significant, and it would be best to directly measure this carving. This is what our significance testing accomplishes. \\

\noindent\paragraph{Wilcoxon Test} To test that answer tokens are more repelling than non-answer tokens we use the Wilcoxon Signed Rank Test (Wilcoxon), and the Rank Biserial Correlation (RBC) to quantify effect size. The Wilcoxon test returns us $W^+, W^-$ which is the sum of signed ranks for the positive and negative differences, where here $W^+$ is the effect size and we expect it to be low (the null hypothesis being that they are equal). We can quantify the difference by taking the RBC given by $\frac{W^+ - W^-}{W^+ + W^-}$ to see how asymmetrical they are. The RBC is bounded between $[-1, 1]$, and we hope to observe a negative correlation, which we do.

\section{Function and Content Separation Figures}
\label{appdx:linguistic_striation}

See the figures below, \autoref{fig:gpt0} to \autoref{fig:gemma1} for more linguistic striation and model processing examples. The layers are chosen arbitrarily, roughly evenly spaced to demonstrate transitions throughout the model.

\begin{figure*}[h!]
\centering
  \includegraphics[width=\textwidth]{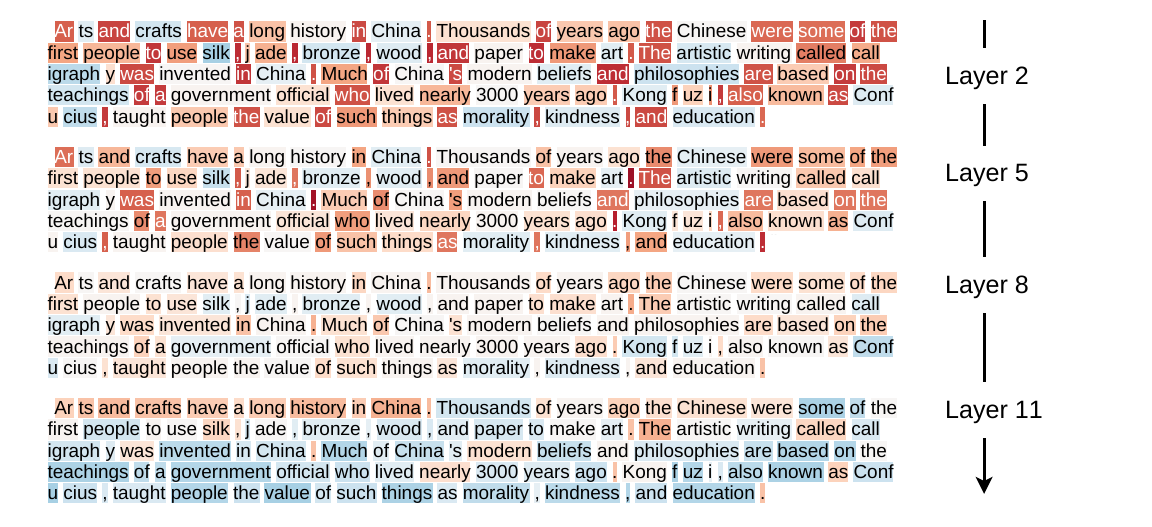}
  \caption{Example of GPT2}
  \label{fig:gpt0}
\end{figure*}

\begin{figure*}[h!]
\centering
  \includegraphics[width=\textwidth]{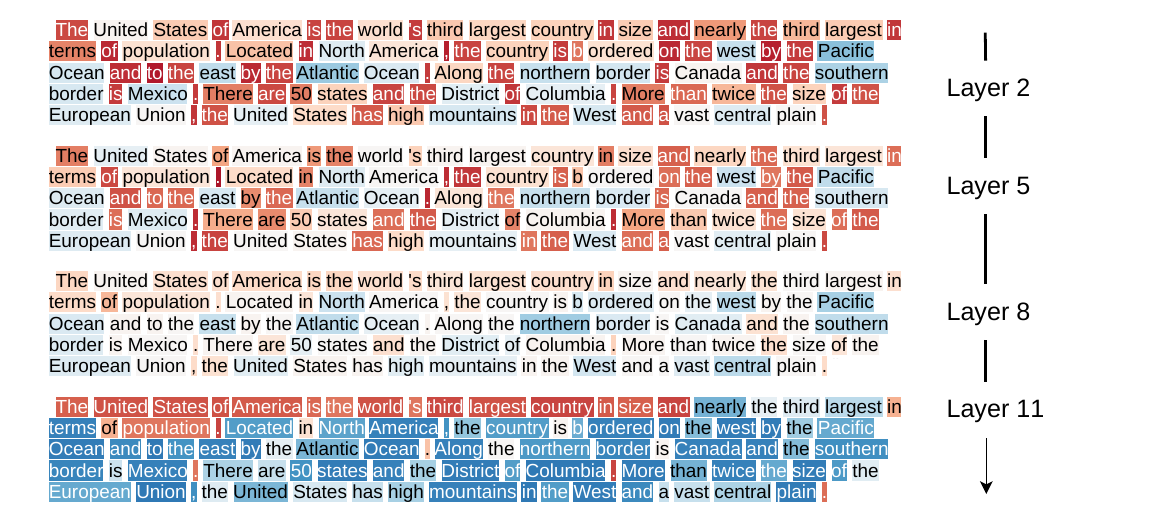}
  \caption{Example of GPT2}
  \label{fig:gpt1}
\end{figure*}

\begin{figure*}[h!]
\centering
  \includegraphics[width=\textwidth]{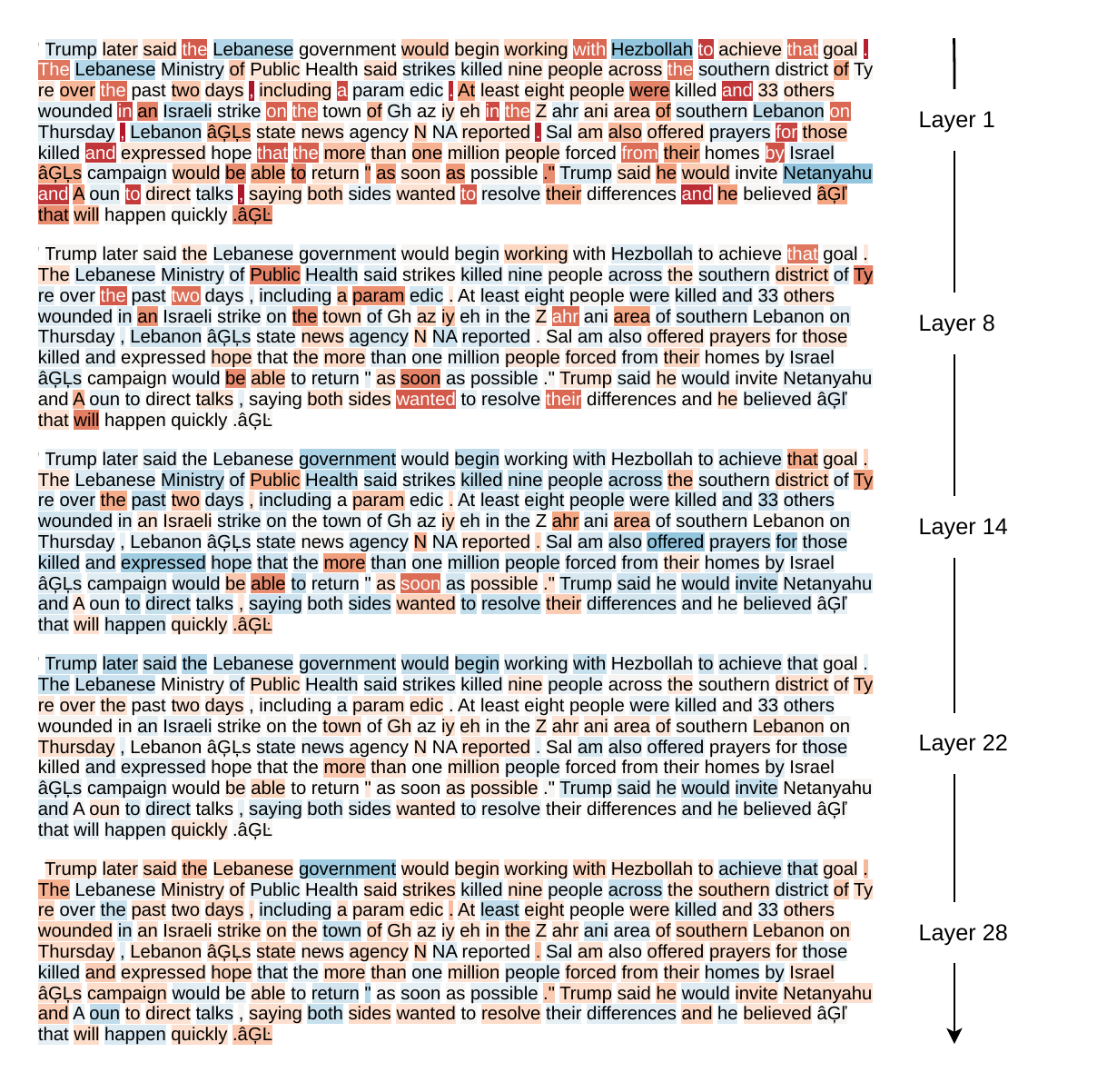}
  \caption{Example of Llama3.2}
  \label{fig:llama0}
\end{figure*}

\begin{figure*}[h!]
\centering
  \includegraphics[width=\textwidth]{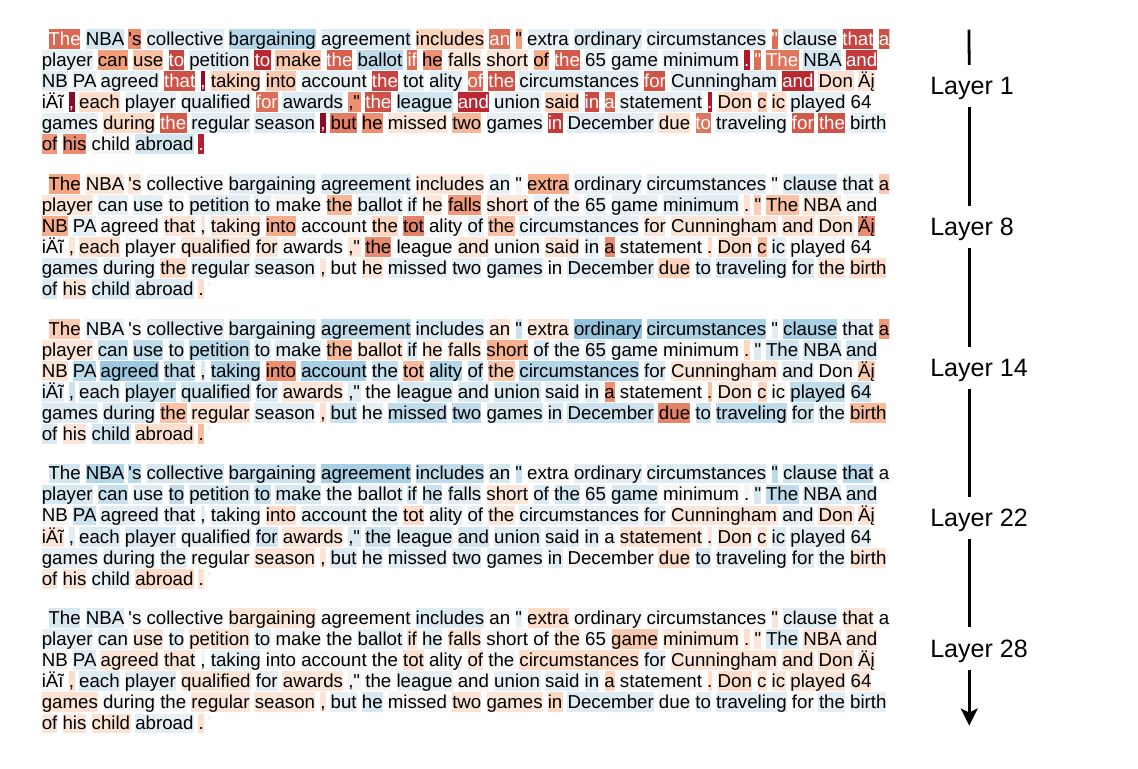}
  \caption{Example of Llama3.2}
  \label{fig:llama1}
\end{figure*}

\begin{figure*}[h!]
\centering
  \includegraphics[width=\textwidth]{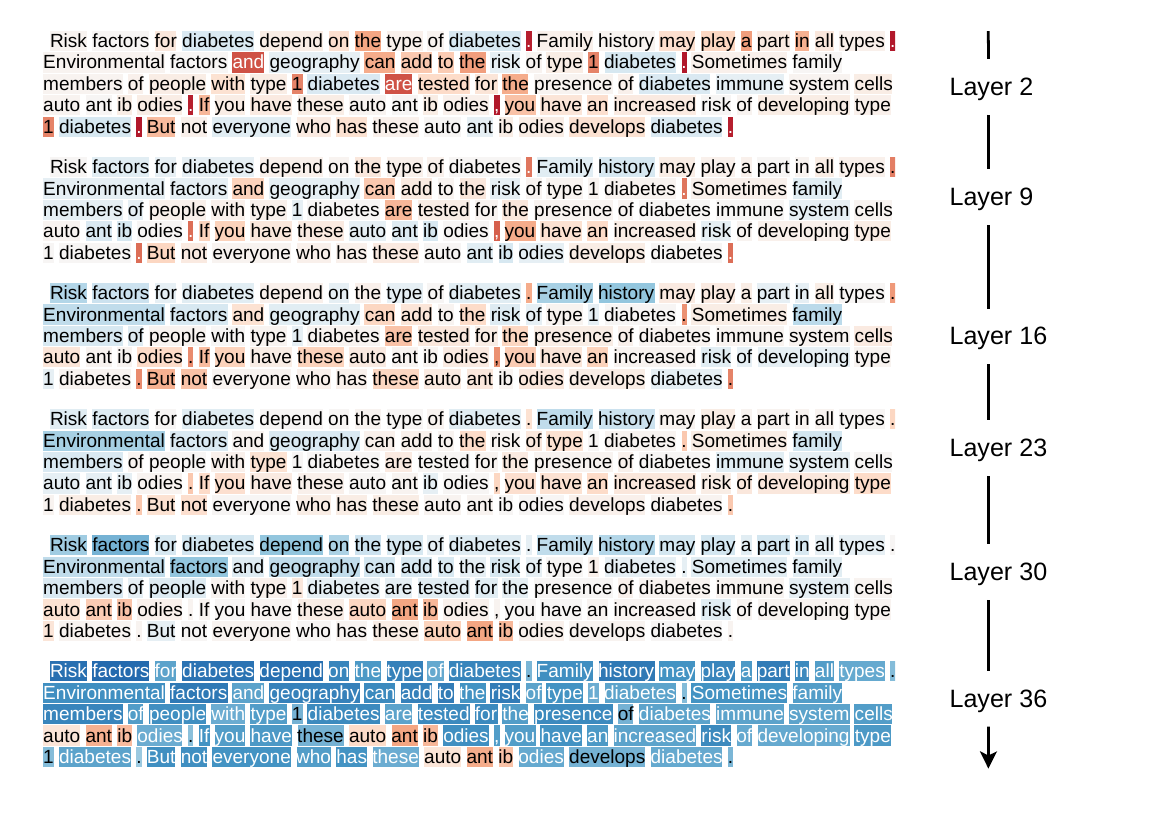}
  \caption{Example of Qwen2.5}
  \label{fig:qwen0}
\end{figure*}

\begin{figure*}[h!]
\centering
  \includegraphics[width=\textwidth]{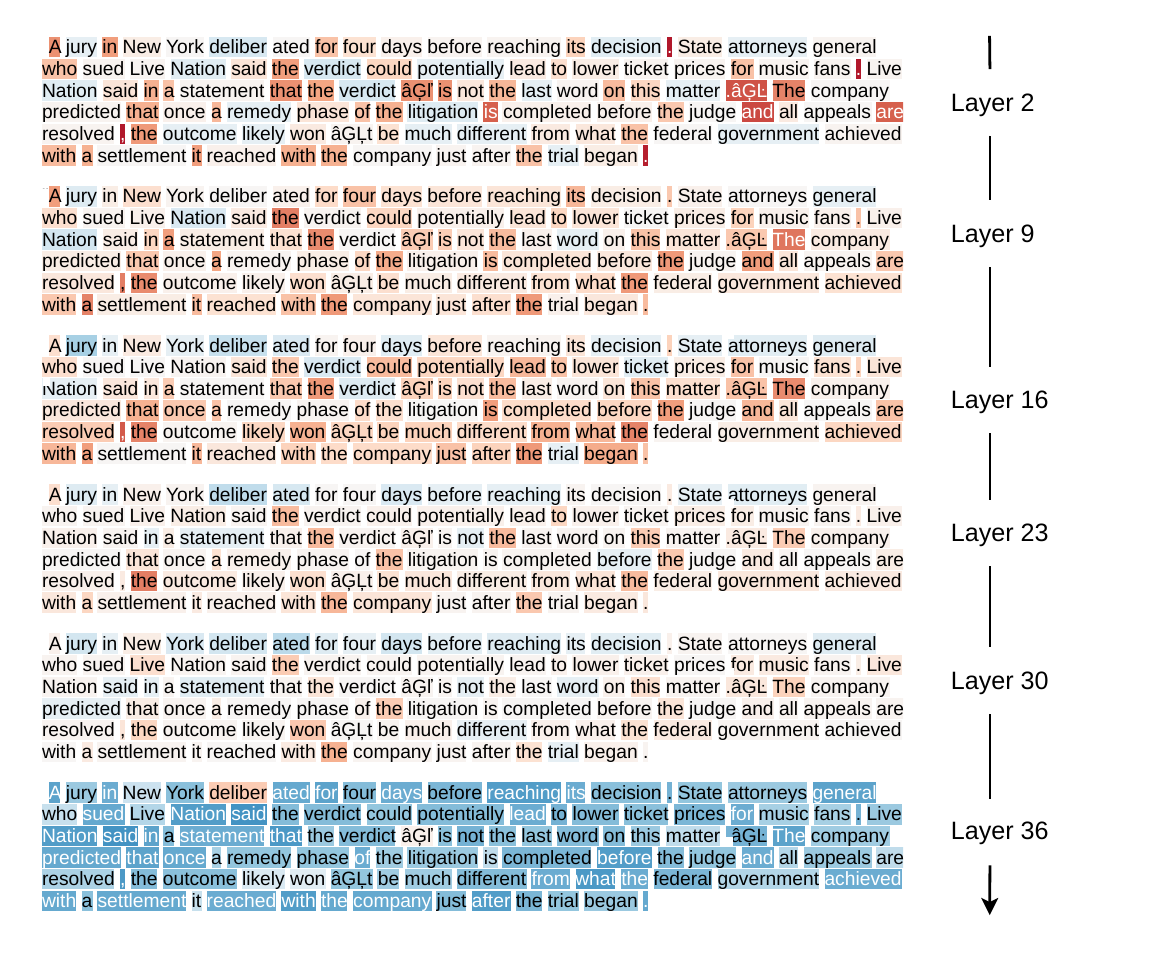}
  \caption{Example of Qwen2.5}
  \label{fig:qwen1}
\end{figure*}

\begin{figure*}[h!]
\centering
  \includegraphics[width=\textwidth]{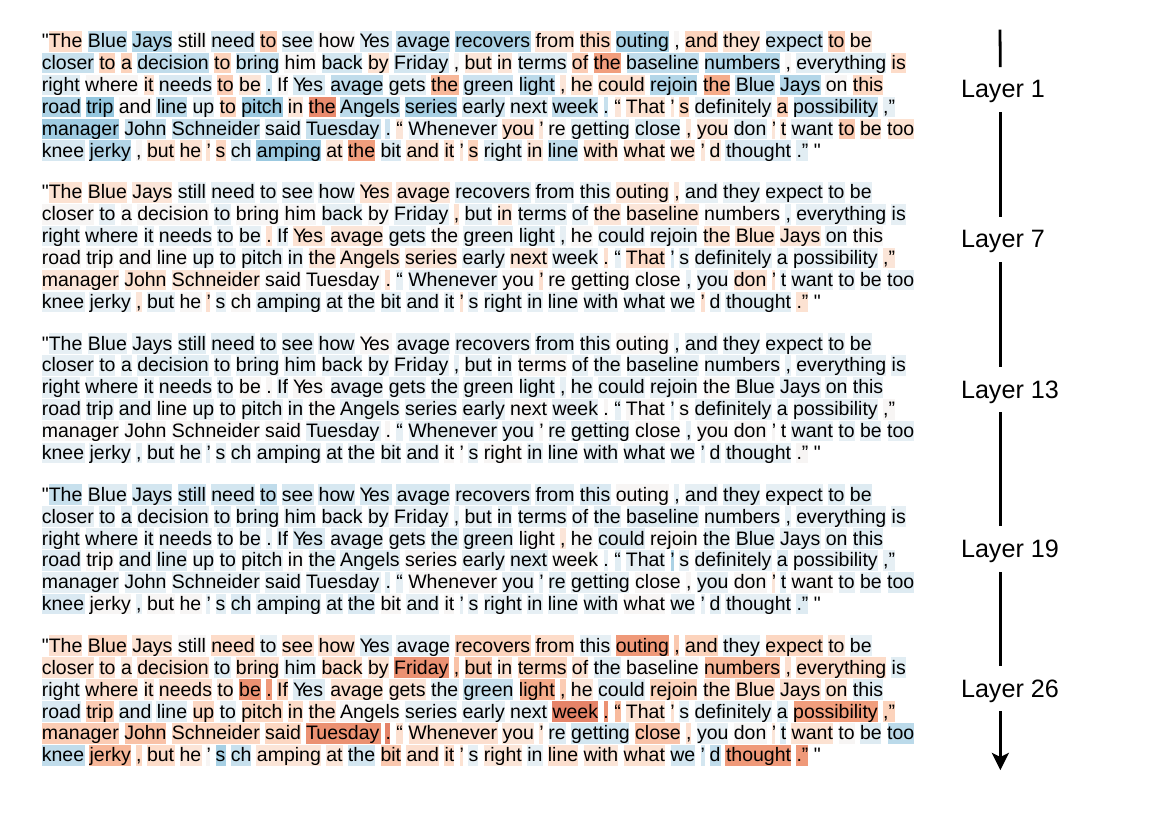}
  \caption{Example of Gemma3}
  \label{fig:gemma0}
\end{figure*}

\begin{figure*}[h!]
\centering
  \includegraphics[width=\textwidth]{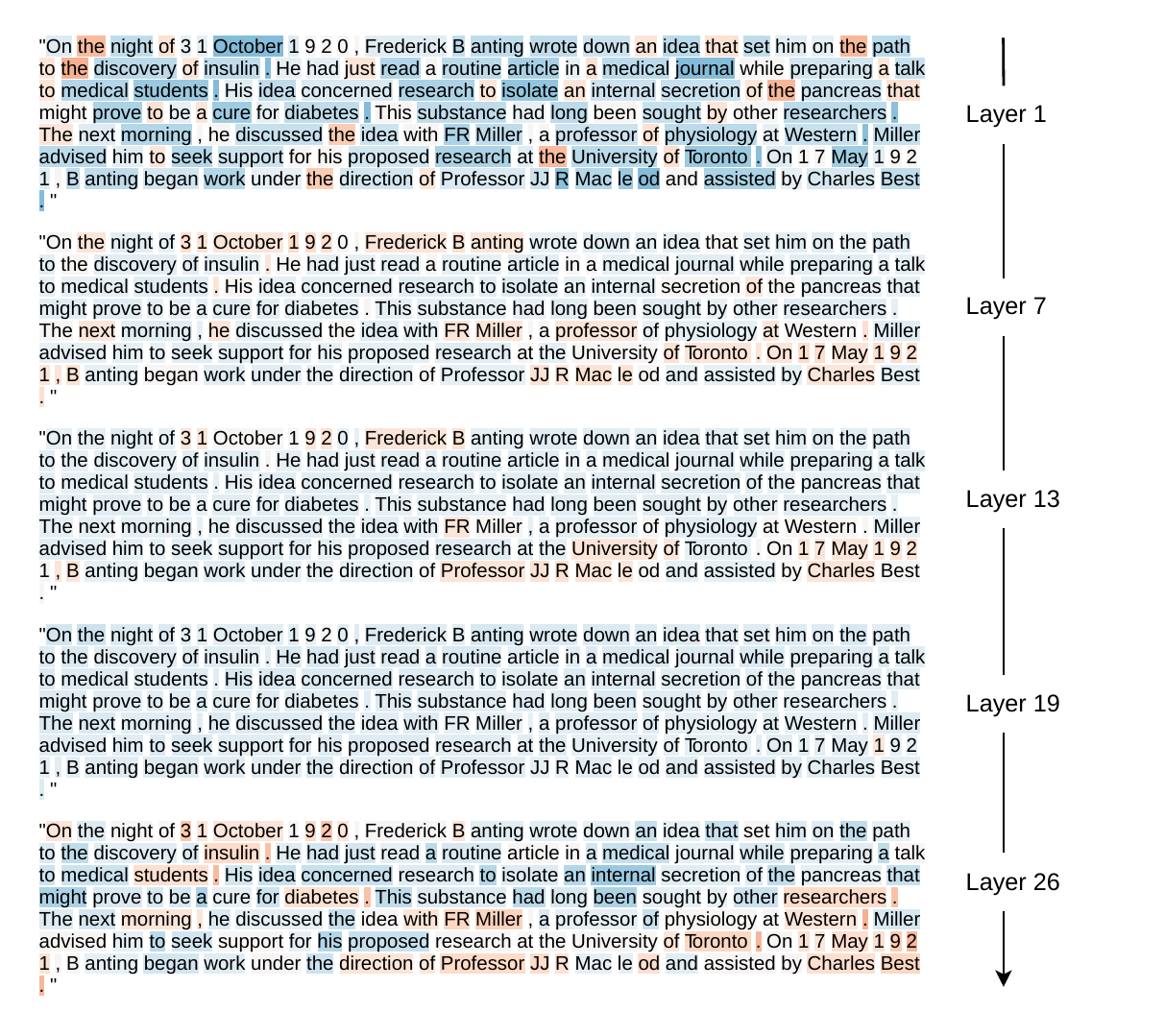}
  \caption{Example of Gemma3}
  \label{fig:gemma1}
\end{figure*}

\end{document}